\RequirePackage{silence}
\documentclass[10pt,twocolumn,letterpaper]{article}

\usepackage{dsfont}
\usepackage{cvpr} %

\usepackage[table, dvipsnames]{xcolor}
\usepackage{graphicx}
\usepackage{adjustbox}
\usepackage{amsmath}
\usepackage{array}
\usepackage{amssymb}
\usepackage{booktabs}
\usepackage{pifont}
\usepackage{pgfplots}
\usepackage{multirow}
\usepackage{tabularx}
\usepackage{comment}
\usepackage{bbm}
\usepackage{ulem}
\usepackage{float}
\usepackage[accsupp]{axessibility}
\newcommand{\whitebox}[1]{\textcolor{black}{\fcolorbox{black}{white}{#1}}}
\newcommand{\inlineColorbox}[2]{\begingroup\setlength{\fboxsep}{1pt}\colorbox{#1}{\hspace{2pt}\vphantom{Ay}#2\hspace{2pt}}\endgroup}

\definecolor{cvprblue}{rgb}{0.21,0.49,0.74}
\usepackage[pagebackref,breaklinks,colorlinks,citecolor=cvprblue]{hyperref}

\definecolor{light}{rgb}{0.68, 0.90, 0.77}
\definecolor{orange}{rgb}{0.93, 0.74, 0.60}
\definecolor{lightorange}{rgb}{1, 0.87, 0.68}
\definecolor{lightgreen}{rgb}{0.76, 0.88, 0.76}
\definecolor{lightgray}{rgb}{0.92, 0.92, 0.92}
\definecolor{lightred}{rgb}{0.92, 0.29, 0.36}
\definecolor{multiflowcolor}{rgb}{0.92, 0.88, 1}
\definecolor{lightcyan}{rgb}{0.424, 0.651, 0.804}

\newcommand{\myparagraph}[1]{\vspace{2pt}\noindent\textbf{#1}}

\newcommand\blfootnote[1]{%
  \begingroup
  \renewcommand\thefootnote{}\footnote{#1}%
  \addtocounter{footnote}{-1}%
  \endgroup
}

\newcommand{\method}{\textsc{multiflow}\xspace}
\newcommand{\methodFull}{Multimodal Flow Pruning\xspace}
\newcommand{\taskFull}{Task-Agnostic Vision-Language Pruning\xspace}
\newcommand{\task}{TA-VLP\xspace}
\newcommand{\xmark}{\ding{55}}

\DeclareMathOperator*{\argmax}{arg\,max}

\def\paperID{15677} %
\def\confName{CVPR}
\def\confYear{2024}

\title{\method: Shifting Towards Task-Agnostic Vision-Language Pruning}

\author{Matteo Farina\textsuperscript{$1$} \quad Massimiliano Mancini\textsuperscript{$1$} \quad
Elia Cunegatti\textsuperscript{$1$} \\
Gaowen Liu\textsuperscript{$2$} \quad
Giovanni Iacca\textsuperscript{$1$} \quad
Elisa Ricci\textsuperscript{$1,3$} \\
\small
$^1$University of Trento \quad $^2$Cisco Research \quad $^3$Fondazione Bruno Kessler \\
}
\pgfplotsset{compat=1.18}

\begin{document}
\maketitle

\begin{abstract}
While excellent in transfer learning, Vision-Language models (VLMs) come with high computational costs due to their large number of parameters. 
To address this issue, removing parameters via model pruning is a viable solution. 
However, existing techniques for VLMs are task-specific, and thus require pruning the network from scratch for each new task of interest. 
In this work, we explore a new direction: {\taskFull (\task)}. 
Given a pretrained VLM, the goal is to find a unique pruned counterpart transferable to multiple unknown downstream tasks. 
In this challenging setting, the transferable representations {already} encoded in the pretrained model are a key aspect to preserve. 
Thus, we propose {\methodFull} (\method), a first, gradient-free, pruning framework for \task where: (i) the importance of a parameter is expressed  in terms of its magnitude and its 
information flow, by incorporating the saliency of the neurons it connects; and (ii) pruning is driven by the emergent (multimodal) distribution of the VLM parameters after pretraining. 
We benchmark eight state-of-the-art pruning algorithms in the context of \task, experimenting with two VLMs, three vision-language tasks, and three pruning ratios.
Our experimental results show that \method outperforms recent sophisticated, combinatorial competitors in the vast majority of the cases,  paving the way towards addressing 
\task.
The code is publicly available at {\small \url{https://github.com/FarinaMatteo/multiflow}}. \blfootnote{Corresponding author: {\tt m.farina@unitn.it}.}
\end{abstract}

\begin{figure}[!ht]
\centering
\includegraphics[width=\columnwidth]{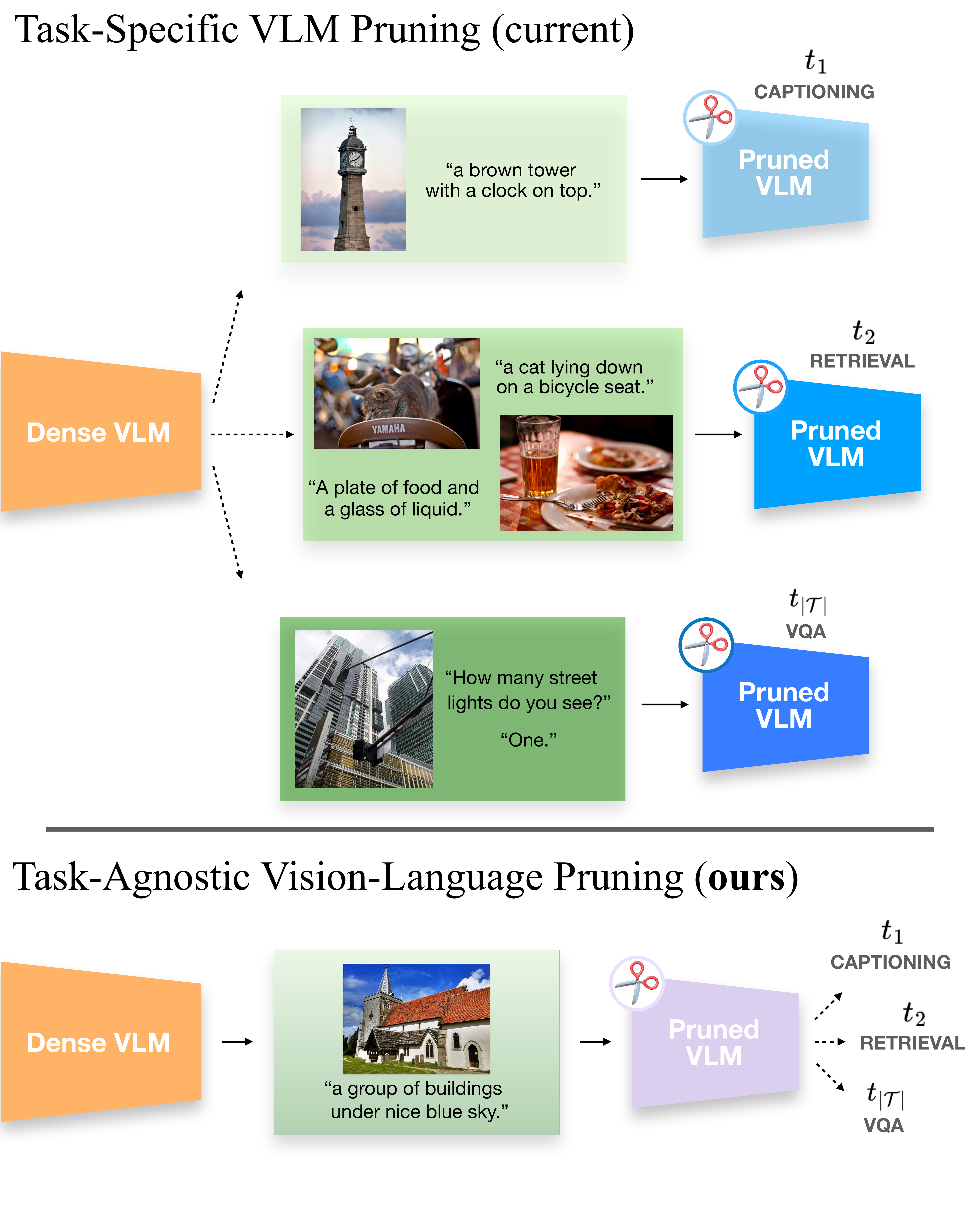}
\caption{The conceptual difference between existing VLM pruning methods \cite{shi2023upop,wang2022efficientvlm} and our proposed \taskFull. While existing pruning methods use task-specific knowledge, hence requiring pruning the dense model from scratch for different tasks, we propose to shift the perspective and formalize \task, which only requires pruning once.}
\label{fig:teaser}
\vspace{-10pt}
\end{figure}

\section{Introduction}
\label{sec:intro}
Large-scale vision-language models (VLMs) \cite{li2022blip,radford2021learning,li2021align,li2023blip2} show remarkable transfer learning capabilities and achieve state-of-the-art results in multiple vision-language tasks after fine-tuning with task-specific data and little architectural changes.
However, these practical advantages come at the price of a huge number of parameters, \eg, in the order of hundreds of millions \cite{li2022blip}, hindering deployment in memory-constrained devices. 
A solution to this problem is to reduce the network size via pruning \cite{gan2022playing}, a compression technique whose aim is to explicitly remove network parameters. 
In the context of VLMs, existing approaches perform pruning for specific downstream tasks \cite{shi2023upop,wang2022efficientvlm}, where the obtained pruned models achieve good transfer learning performance once fine-tuned.
Critically, this requires re-pruning the model from scratch if the downstream task changes. 
This is not only time-consuming, but further assumes that the original VLM parameters can easily be stored. 
To overcome these issues, we tackle the problem from a different perspective, investigating the possibility of pruning a VLM while maintaining its general transfer learning capabilities. 
 Towards this goal, we propose \taskFull (\task), where the aim is to prune a VLM \textit{once} and obtain a sparse model transferable to multiple \textit{unknown} tasks when fine-tuned (see 
 Fig.~\ref{fig:teaser}). 
 While appealing, finding an optimal solution to \task is challenging as we cannot use any task-specific priors nor feedback during pruning, and different downstream tasks may focus on different visual/linguistic cues (\eg, local for visual question answering, global for captioning).
 Nevertheless, we can still rely on one anchor: the pretrained VLM. 
 In fact, pretraining uses a generic objective, such as vision-language alignment, which applied to large-scale data enables learning generic and transferable representations. 
 These representations depend on the network parameters and on how the (multimodal) activations propagate through the network. 
 Intuitively, if we assume the pretrained model to be transferable, its pruned counterpart should preserve the learned activation patterns. 
 
 Following this principle, we propose \methodFull (\method), a first method for \task. 
 \method models each layer as a bipartite graph, where nodes are activations and edges are parameters. 
 Exploiting calibration data, the saliency of a parameter is modeled by combining its magnitude with the average signal emitted/aggregated by the input/output nodes it connects. 
However, directly pruning using these scores may lead to biases w.r.t. the depth of a layer, and ignores that activation patterns and magnitudes may differ among modalities. 
To overcome this issue, we disentangle modalities and guide pruning with the emergent distribution of the magnitude of the parameters. 
Our experiments on XVLM \cite{zeng2022multi} and BLIP \cite{li2022blip}, with three vision-language tasks and pruning ratios, show that \method consistently matches or surpasses existing methods while requiring no gradient information. 

\myparagraph{Contributions.} To summarize, our contributions are:
 \begin{itemize}
 \item We formalize \taskFull, whose aim is to prune a VLM \textit{once} while maintaining transferability to \textit{unknown} downstream tasks. 
 \item We propose \methodFull, a first specific method for \task, where the importance of a parameter depends on the aggregated importance of the nodes it connects and its magnitude, exploiting multimodal priors to guide the distribution of each layer and avoid biases.
 \item We benchmark existing methods and \method on \task, with multiple vision-language tasks, VLMs, and pruning ratios, demonstrating the effectiveness of \method. 
 These results and the proposed benchmark also highlight a large gap \wrt the performance of the dense model, paving the way for future research on this topic.
 \end{itemize}

\section{Related Work}
\label{sec:rw}

\myparagraph{Post-training pruning.} Several works remove parameters after training \cite{lecunoptimal,hassibisecond,han2015deep,han2015learning,dong2017learning,lee2021layeradaptive}, aiming to reduce inference time \cite{molchanov2017pruning,molchanov2019importance} and storage requirements \cite{zhu2017prune,gale2019state}. While some of these techniques are data-free and mostly rely on weight magnitude, others are data-driven and exploit first- or second-order information. 
Another line of data-driven algorithms relies on combinatorial optimization \cite{singh2020woodfisher,yu2022combinatorial,benbaki2023fast} or iterative procedures, like iterative magnitude pruning (IMP) \cite{frankle2018the, Renda2020Comparing}, which alternates training until convergence, magnitude pruning, and weight rewinding.  
While we also exploit the weight magnitude in our pruning criterion, we additionally consider how information propagates through the target VLMs, fostering transferability to \textit{unknown} tasks. 

\myparagraph{Pruning at initialization.} Since our goal is to prune the model \emph{before} fine-tuning it on downstream tasks, our work is also closely related to pruning at initialization (PaI).
PaI methods rely on a \textit{saliency} function that evaluates each connection, removing those with the lowest saliency scores to meet a target sparsity level \cite{lee2018, wang2020, alizadeh2022prospect}.
Notably, the same saliency scores can be re-used for different sparsity targets 
without any additional computational overhead. 
The saliency function can also be applied iteratively, trading-off efficiency for performance \cite{Jorge2020ProgressiveST,Tanaka2020PruningNN,wang2023ntksap}.
 PaI methods can be categorized based on their input as well, with data-dependent algorithms \cite{lee2018,Jorge2020ProgressiveST,alizadeh2022prospect,wang2020} employing samples from the target dataset, and data-free techniques using synthetic inputs  \cite{Tanaka2020PruningNN, wang2023ntksap}. 
 Our work shares the same rationale of these techniques, \ie, the saliency of network connections is central to our pruning strategy.
 However, while existing PaI methods mostly focus on task-specific objectives with backward gradient propagation, \method is gradient-free, as it is only based on the forward function of the model. 
 
\myparagraph{Pruning in Vision and/or Language.} Several works explore pruning in the context of vision and language, with works sparsifying large pretrained models in NLP, \eg, BERT \cite{zafrir2021prune,kurtic2022optimal, kwon2022a,jaiswal2023emergence}, GPT \cite{frantar2023sparsegpt} and LLaMA \cite{sun2023simple}, and vision transformers \cite{dosovitskiy2020image}, \eg, via model \cite{lee2021layeradaptive,zhu2021vision,jaiswal2023emergence,yang2023global} or token~\cite{rao2021dynamicvit,liang2022evit,meng2022adavit,tang2022patch,bonnaerens2023learned} pruning. 
To the best of our knowledge, UPop \cite{shi2023upop} and EfficientVLM \cite{wang2022efficientvlm} are the only pruning algorithms specifically developed and benchmarked on multiple vision-language tasks. 
The former progressively prunes the target VLM during fine-tuning, while the latter is a distill-then-prune framework. 
However, both methods require task-specific knowledge \emph{by design}. 
In this work, we take instead a different direction, 
investigating how we can extract task-agnostic subnetworks from VLMs. 

\myparagraph{Unimodal Task-Agnostic Pruning.} 
In the context of Vision- or Language-only, some works already explored the ability of pruned models to generalize to multiple tasks \cite{xu2022dense}. 
For instance, \cite{chen2020lottery, chen2021lottery} assess that the \emph{train-prune-retrain} paradigm of IMP can also be successfully applied during unimodal pretraining.
In principle, a task-agnostic subnetwork can also emerge when pretraining by optimizing the pruning masks as trainable parameters (a common procedure in continual learning \cite{mallya2018piggyback} or neural architecture search \cite{chavan2022vision}), as shown in \cite{liu2022learning}.
However, complete access to the pretraining phase is often out of reach due to its large computational demand. 
Exploratory works also study the effects of pruning on BERT transfer \cite{gordon2020compressing}, highlighting that finetuning recovers dense performance when less than half of the parameters are pruned, or on self-supervised CNNs \cite{caron2020pruning}.

Our %
work has different rationales.
Motivated by \cite{gan2022playing}, which showcases the additional difficulty in finding task-agnostic pruned models with Vision and Language \wrt unimodal scenarios, we design the first algorithm for TA-VLP. 
We do not assume access to large amounts of data nor the pretraining phase, striving for \emph{fast} and \emph{efficient} pruning.
Hence, we avoid the burden of both computing and storing gradients for large VLMs with a gradient-free algorithm.
Neither IMP nor Mask Training meet these principles.

\section{\taskFull}
\label{sec:tavlp}
In this section, we formally define the \emph{\taskFull} (\task) problem, discussing its challenges and its relation with prior work on pruning VLMs.

 \myparagraph{Preliminaries.} Let $f$ denote a VLM, and let $\Theta \in \mathbbm{R}^n$ be its corresponding parameters after pretraining on a large-scale dataset $\mathcal{D}_p$ of image-text pairs. 
Given data for a specific vision-language task $t$, we can fine-tune $\Theta$ to improve performance on $t$ itself.
While the standard practice is to directly update $\Theta$, the latter is extremely high-dimensional and cannot always be stored. 
To circumvent this issue, pruning algorithms for VLMs \cite{shi2023upop} prune $\Theta$ explicitly for $t$. 
Formally, given task data $\mathcal{D}_t$, they aim at a binary mask $\mathbf{m}_t \in \mathbbm{B}^n$, by maximizing a task-dependent criterion $\mathcal{C}_t$: 
\begin{equation} \label{eq:ts-vlp}
\begin{split}
 \mathbf{m}_t = \argmax_{\mathbf{m}}\; 
 &\mathcal{C}_t \Bigl(f(\Theta \odot \mathbf{m}), \mathcal{D}_t \Bigr) \\
 &\text{s.t. } ||\mathbf{m}||_0 = k
\end{split}
\end{equation}
where $k$ denotes the sparsity constraint (\ie, the number of parameters to preserve). 
This mask should maximize the performance $\mathtt{p}_t$ of model $f$ on task $t$, when $f$ is trained using a given algorithm $\mathcal{A}_t$, \ie, $\mathtt{p}_t(\mathcal{A}_t, f, \Theta \odot \mathbf{m}_t)$. 
However, with this setup one needs to re-prune the model from scratch for every new task, which requires both time and storage of the original $\Theta$. 
To overcome these issues, we shift the perspective towards task-agnostic model pruning. 

\myparagraph{Task-agnostic VLM Pruning.} The goal of \task is to prune a VLM once while preserving trainability for arbitrary downstream tasks, without re-compressing the model from scratch. 
Formally, we aim for a task-agnostic mask $\mathbf{m}_a$ that maximizes the performance of $f$ on a series of \textit{unknown} downstream tasks $\mathcal{T}$:
\begin{equation} \label{eq: tavlp}
 \sum_{t\in \mathcal{T}} \mathtt{p}_t (\mathcal{A}_t, f, \Theta \odot \mathbf{m}_a). 
\end{equation}
As it is unfeasible to collect data for \textit{unknown} target tasks, a \task algorithm should produce $\mathbf{m}_a$ %
from a generic dataset $\mathcal{D}_g$ and a generic criterion $\mathcal{C}_g$, \ie:
\begin{equation} \label{eq:ta-vlp}
\begin{split}
 \mathbf{m}_a = \argmax_{\mathbf{m}}\; 
 &\mathcal{C}_g \Bigl(f(\Theta \odot \mathbf{m}), \mathcal{D}_g \Bigr) \\
 &\text{s.t. } ||\mathbf{m}||_0 = k.
\end{split}
\end{equation}
In this work, we always assume that $\mathcal{D}_g$ is much smaller than the pretraining dataset $\mathcal{D}_p$, \ie, $|\mathcal{D}_g|<<|\mathcal{D}_p|$, as the aim of \task should not be to re-train a smaller model from scratch, but rather to efficiently prune an existing VLM.
\task entails several challenges, as different tasks may exhibit stronger sensitivity to visual, textual, or fused knowledge.  
Intuitively, the solution to \task requires finding the optimal trade-off among modalities and encoded knowledge, respecting the priors of the pretrained VLM. 
In the following, we describe how we tackle \task by considering the multimodal information flow within VLMs. 

\section{\method: Multimodal Flow Pruning}
\label{sec:mfp}
In this section we introduce \emph{\textbf{Multi}modal \textbf{Flow} Pruning}, a first algorithm for (unstructured) \taskFull. %
We first discuss how we model the information flow within the VLM, and then how we exploit (multimodal) pretraining priors when pruning. %

\subsection{Modeling the Information Flow}
As we lack task priors, we have one anchor when performing \task: the pretrained VLM. 
In fact, if we assume that the pretrained VLM encodes transferable representations, preserving how these representations emerge should maintain also the transferability to downstream tasks.  
We exploit this principle and tackle the problem from the perspective of the information flowing through the network, %
framing network pruning through the lens of message passing. %

Without loss of generality, let us focus on a linear layer. %
We can represent a dense linear projection as a directed, weighted and complete bipartite graph ${G} = ({L} \cup {R}, {E})$, where ${L} $ and ${R}$ are disjoint sets of nodes and ${E}$ %
is the set of edges connecting them. Note that an edge $e_{lr}\in{E}$ connecting nodes $l\in L$ and $r \in R$ corresponds to a parameter $\theta_{lr}\in \Theta$. In this context, we model the importance of $\theta_{lr}$ as the %
information passing through its corresponding edge $e_{lr}$. %
This  %
depends on three different values: (i) the weight of the edge, (ii) the saliency of the input node $l$, and (iii) the saliency of the output node $r$. 
In the following, we describe how we measure and combine these components.  

\begin{figure*}[!ht]
\centering
\includegraphics[width=2\columnwidth]{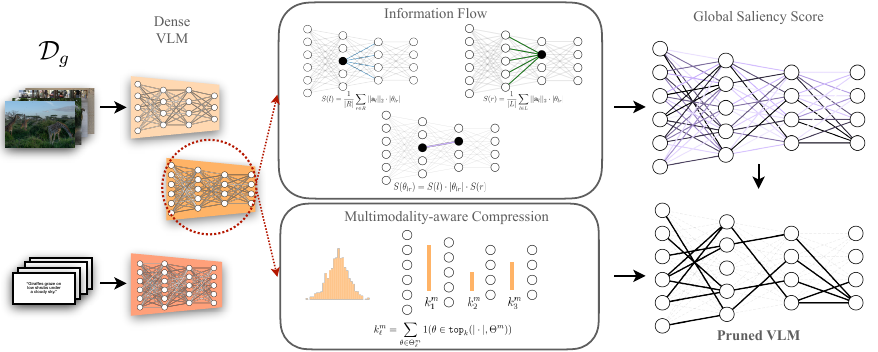}

\caption{\textbf{\method}. Orange trapezoids represent groups of parameters processing different \inlineColorbox{orange}{modalities} (i) To compute the \textbf{information flow} score for a parameter $\theta_{lr}$, \method combines the importance of the \inlineColorbox{lightcyan}{input neuron} $l$ and that of \inlineColorbox{lightgreen}{output neuron} $r$, aggregating them via the \inlineColorbox{multiflowcolor}{local hop} from $l$ to $r$ through $\theta_{lr}$ (ii) A \textbf{global saliency score} is obtained by computing (i) for all edges, and a \textbf{global modality-aware distribution} that exploits the emergent properties of large-scale pretraining guides layer-wise pruning.}
\label{fig:method}
\end{figure*}

\myparagraph{Importance of an edge.} This component is directly estimated from the absolute value of its corresponding weight, \ie, $I(e_{lr}) = |\theta_{lr}|$, a simple and effective importance estimation technique well-known in the literature \cite{han2015deep,frankle2018the}. 
Note that this measure inherently exploits pretraining information, as weight magnitudes can be viewed as long-term accumulators of movement in the parameter space as an emergent property from pretraining \cite{sanh2020movement}. 

\myparagraph{Saliency of a node.} 
In each layer, input and output nodes have distinct roles: the former account for forwarding information, while the latter account for aggregating it.
Given a vector $\mathbf{a}_l \in \mathbbm{R}^{N}$ collecting $N$ activations to an input node $l$ of a given layer, we can anchor on the distinct roles of the neurons to assign ad-hoc saliency criteria.
Following this intuition, we frame the saliency of $l \in L$ as the average strength of the signal it \emph{emits} towards \emph{all} output neurons: %
\begin{equation} \label{eq:left}
 S(l) = \frac{1}{|{R}|}\sum_{r\in R} ||\mathbf{a}_l||_2 \cdot |\theta_{lr}|.
\end{equation} 
On the other hand, we treat the saliency of an output node $r \in R$ as the average strength of the signals it \emph{receives}:
\begin{equation} \label{eq:right}
 S(r) = \frac{1}{|{L}|}\sum_{l \in L} ||\mathbf{a}_l||_2 \cdot |\theta_{lr}|. 
\end{equation}
Note that we use magnitude and norms to avoid potential sign misalignments between the two: those would create misleading importance scores as, \eg, a negative weight with large magnitude may greatly influence the output and vice versa. 
To ensure that Eqs.~\eqref{eq:left} and ~\eqref{eq:right} take into account both the modulation of the edges and the activation patterns grounded in the VLM forward process, we estimate the activation norms over the available calibration data $\mathcal{D}_g$, similarly to concurrent work on LLM pruning \cite{sun2023simple}.
Notably, this only requires forwarding $\mathcal{D}_g$ through the model, which is much faster than also computing gradients.

\myparagraph{Final score.} Collating previous concepts, we define the saliency of each parameter $\theta_{lr}$ connecting nodes $l$ and $r$ via the edge $e_{lr}$, as the saliency of the path from $l$ to $r$:
\begin{equation} \label{eq:final_score}
 S(\theta_{lr}) = S(l)\cdot I(e_{lr}) \cdot S(r) = S(l)\cdot |\theta_{lr}| \cdot S(r). 
\end{equation}
Note that Eq.~\eqref{eq:final_score} equally balances the contribution of each graph part, involving both the information captured by the edge weights and its relation to the most salient nodes. 

\subsection{Multimodality-aware compression} \label{sec:mmp}Properly defining a saliency criterion ultimately enables ranking the parameters of a model and, consequently, network pruning. 
Given a target sparsity constraint $k$, the most straightforward solution would be to instantiate a binary mask preserving the top-k parameters according to the scores in Eq.~\eqref{eq:final_score}. 
However, this would ignore potential sources of bias. 
For instance, as deeper layers accumulate magnitude from preceding ones, this may cause a large discrepancy in the scores, with the risk that the pruning criterion penalizes early layers and induces layer collapse \cite{Tanaka2020PruningNN}. 
This also applies to the multimodal nature of the model: as different layers may receive inputs from different modalities,  
by assuming that information equally flows among them, we may overlook their respective distribution, hence biasing our pruning mask on one of them. We provide more insights on both these phenomena in Sec. \ref{sec:abl}.

To avoid biasing the model towards a specific modality and/or network level, we re-weight the importance of each parameter based on the prior distribution given by the pretrained VLM parameters. 
In fact, we found that the magnitude of the weights \emph{detached} from the input/output information flow is a good indicator of the overall distribution that the pruned network should maintain layer-wise. 
This estimation tends to be more accurate if we keep into account which modality the layer processes, and its relevance increases with the disentanglement among modalities.
Formally, given a layer $\ell$ processing information from modality $m$ (\eg, visual or textual), with its corresponding parameters $\Theta^m_\ell$, we define its active parameter count $k_\ell^m$ as:
\begin{equation} \label{eq:sparsity}
 k^m_\ell = \sum_{\theta \in \Theta_\ell^m} \mathds{1}(\theta \in \mathtt{top}_k(|\cdot|, \Theta^m))
\end{equation}
where $\mathds{1}$ is the indicator function and $\Theta^m\subset \Theta$ is the subset of parameters processing a specific modality $m$ as input. $\mathtt{top}_k(|\cdot|, \Theta^m)$ is the set of top-k elements in $\Theta^m$ if we rank elements according to their magnitude.
Given this prior distribution $k^m_\ell$, the final mask for the layer is:
\begin{equation}\label{eq:final-mask}
 \mathbf{m}^{\ell,m}_{lr} = \begin{cases}
 1\; \text{if}\; \theta_{lr} \in \mathtt{top}_{k^m_\ell}(S,\Theta^m_\ell)\\
 0\; \text{otherwise}
\end{cases} 
\end{equation}
using $S$ as criterion for the $\mathtt{top}_k$. With Eq.~\eqref{eq:final-mask}, the compressed network will take into account: (i) the information flow (as estimated via Eq.~\eqref{eq:final_score}), (ii) the bias of the information at different levels of the network, and (iii) the peculiar flow of each modality (Eq.~\eqref{eq:sparsity}). 
All this information, combined, allows the pruned model to maintain priors of the original model parameters 
and preserve core connections.
The overall procedure of \method is depicted by Fig. \ref{fig:method}.

\section{Experiments} \label{sec:experiments}
In this section, we benchmark well-established and 
recent pruning algorithms in \task.
We experiment with three different downstream tasks: Image-Text Retrieval, Image-Captioning, and Visual Question Answering \cite{lin2014microsoft, chen2015microsoft, goyal2017making}. 
We additionally report Vision-only experiments in Sec.~\ref{sec:dino-exps}.

\myparagraph{Architectures.} To study how distinct model designs impact \task, we experiment with two significantly different VLMs: BLIP\textsubscript{BASE} \cite{li2022blip}, which uses a multimodal mixture of encoder-decoder networks, and XVLM\textsubscript{CLIP}\cite{zeng2022multi}, where a vision-encoder and a text-encoder process information in parallel streams, with a final fusion encoder merging their output.
While BLIP's design is unique, many modern VLMs share the principles of modality separation (\emph{e.g.}, \cite{radford2021learning}) and fusion (\emph{e.g.}, \cite{li2021align,li2023blip2}) at the core of XVLM.
For both, their vision encoders use 16$\times$16-sized patches.

\begin{table*}[!ht]
 \centering
 \adjustbox{width=2\columnwidth}{
 {\scriptsize
 \begin{tabularx}{2\columnwidth}{
 >{\raggedright\arraybackslash}p{1.2cm}
 >{\centering\arraybackslash}p{1.2cm}
 >{\centering\arraybackslash}X
 >{\centering\arraybackslash}X
 >{\centering\arraybackslash}X
 >{\centering\arraybackslash}X
 >{\centering\arraybackslash}X
 >{\centering\arraybackslash}X
 >{\centering\arraybackslash}X
 >{\centering\arraybackslash}X
}

 \toprule
 \multirow{4}{*}{Method} & \multirow{4}{*}{Sparsity} & \multicolumn{4}{c}{BLIP\textsubscript{BASE}} & \multicolumn{4}{c}{XVLM\textsubscript{CLIP}} \\
 \cmidrule(lr){3-6}
 \cmidrule(lr){7-10}
 & & \multicolumn{2}{c}{Image-to-Text [\%]} & \multicolumn{2}{c}{Text-to-Image [\%]} & \multicolumn{2}{c}{Image-to-Text [\%]} & \multicolumn{2}{c}{Text-to-Image [\%]} \\

 \cmidrule(lr){3-4}
 \cmidrule(lr){5-6}
 \cmidrule(lr){7-8}
 \cmidrule(lr){9-10}
 & & R@1 & R@5 & R@1 & R@5 & R@1 & R@5 & R@1 & R@5 \\ 
 \midrule

 \textsc{dense} & $0\%$ & \cellcolor{lightgreen}80.72 & \cellcolor{lightgreen}95.08 & \cellcolor{lightgreen}62.99 & \cellcolor{lightgreen}85.27 & \cellcolor{lightorange}78.18 & \cellcolor{lightorange}94.48 & \cellcolor{lightorange}60.94 & \cellcolor{lightorange}84.41 \\

 \textsc{random} & \multirow{8}{*}{$63\%$}& 0.02 & 0.12&0.02& 0.10 &21.48 & 46.66 & 14.89 & 37.61 \\
 \textsc{snip} & & 68.06$\pm$0.36& 89.63$\pm$0.06& 51.85$\pm$0.12& 78.61$\pm$0.03& 70.19$\pm$0.15 & 91.27$\pm$0.17 & 53.48$\pm$0.11 & 80.22$\pm$0.04 \\
 \textsc{itersnip}\textsuperscript{$\ast$} && 0.03$\pm$0.01 & 0.14$\pm$0.04 & 0.02$\pm$0.00 & 0.10$\pm$0.01 & 67.12$\pm$1.33 & 89.18$\pm$0.94 & 50.78$\pm$0.89 & 78.63$\pm$0.60 \\
 \textsc{omp} & &\underline{75.39$\pm$0.24}&\underline{92.95$\pm$0.18} & \underline{58.71$\pm$0.22}& \underline{82.82$\pm$0.08}& 76.02$\pm$0.64 & 93.35$\pm$0.20 & 58.96$\pm$0.02 & 83.26$\pm$0.12 \\
 \textsc{lamp} & & 70.36$\pm$0.12&90.49$\pm$0.20 & 53.83$\pm$0.21& 79.62$\pm$0.16 &75.32$\pm$0.21 & 93.27$\pm$0.13 & 58.38$\pm$0.31 & 83.17$\pm$0.09 \\
 \textsc{chita} & & 74.36$\pm$0.13 & 92.06$\pm$0.25 & 57.44$\pm$0.16& 82.20$\pm$0.17 &76.05$\pm$0.12 & 93.69$\pm$0.04 & 58.98$\pm$0.08 & \underline{83.39$\pm$0.02} \\
 \textsc{chita++} &&75.00$\pm$0.29& 92.59$\pm$0.14 & 58.01$\pm$0.09& 82.29$\pm$0.19 & \underline{76.59$\pm$0.20} & \underline{93.70$\pm$0.27} & \underline{59.31$\pm$0.11} & 83.34$\pm$0.05 \\
 \rowcolor{multiflowcolor}\method & & \textbf{76.31$\pm$0.09}&
 \textbf{93.27$\pm$0.10} & \textbf{59.02$\pm$0.09}&
 \textbf{83.08$\pm$0.10}
 &\textbf{77.35$\pm$0.51} & \textbf{93.75$\pm$0.04} & \textbf{60.21$\pm$0.16} & \textbf{83.99$\pm$0.05} \\

 \midrule

 \textsc{random} & \multirow{8}{*}{$75\%$} & 0.06& 0.08 & 0.05& 0.14&12.18 & 30.98 & 8.63 & 24.35 \\
 \textsc{snip} & & 51.33$\pm$0.49&79.51$\pm$0.79 & 37.62$\pm$0.81& 67.08$\pm$0.65 & 57.83$\pm$0.59 & 84.78$\pm$0.52 & 43.10$\pm$0.53 & 72.67$\pm$0.41 \\
 \textsc{itersnip}\textsuperscript{$\ast$} & & 0.03$\pm$0.01 & 0.12$\pm$0.07 & 0.04$\pm$0.03 & 0.14$\pm$0.07 & 46.16$\pm$0.84 & 76.13$\pm$0.89 & 33.52$\pm$0.51 & 64.14$\pm$0.56 \\
 \textsc{omp} & & 63.37$\pm$0.35 & 85.97$\pm$0.58 & 48.28$\pm$0.35 & 75.47$\pm$0.17 & 70.27$\pm$0.28 & 90.91$\pm$0.35 & 53.85$\pm$0.12 & 80.22$\pm$0.22 \\
 \textsc{lamp}\textsuperscript{$\ast$} & & 2.10$\pm$3.57& 5.93$\pm$10.03 & 1.46$\pm$2.49& 4.75$\pm$8.06& 69.38$\pm$0.28 & 90.97$\pm$0.38 & 53.15$\pm$0.43 & 79.96$\pm$0.30 \\
 \textsc{chita}\textsuperscript{$\ast$} && 0.99$\pm$1.63&1.49$\pm$2.29 & 1.02$\pm$1.71& 1.59$\pm$2.55 & 70.15$\pm$0.52 & 91.01$\pm$0.13 & 54.05$\pm$0.09 & 80.36$\pm$0.03 \\
 \textsc{chita++} & & \underline{64.62$\pm$0.26}& \underline{87.07$\pm$0.12} & \underline{48.72$\pm$0.14}& \underline{76.25$\pm$0.13}& \underline{70.33$\pm$0.04} & \underline{91.27$\pm$0.13} & \underline{54.32$\pm$0.17} & \underline{80.61$\pm$0.16} \\
 \rowcolor{multiflowcolor} \method &&\textbf{65.73$\pm$0.60}&\textbf{87.97$\pm$0.52} & \textbf{49.85$\pm$0.59}& \textbf{77.18$\pm$0.45} & \textbf{73.87$\pm$0.13} & \textbf{92.91$\pm$0.23} & \textbf{56.94$\pm$0.10} & \textbf{82.29$\pm$0.07} \\
 \bottomrule
 
 \end{tabularx}
 }
}
 \caption{Results for Image-Text Retrieval on COCO at $63\%$ and $75\%$ sparsity. The dense \colorbox{lightgreen}{\textsc{blip}} and \colorbox{lightorange}{\textsc{xvlm}} upper bounds are reported on top. For further context, we include the random baseline as the lower bound. The \textbf{best performer} is bold; the \underline{second best} is underlined. The superscript \textsuperscript{$\ast$} denotes algorithms that perform comparably to the random baseline when pruning \textsc{blip}.}
 \label{tab:itr}
 \vspace{-10pt}
\end{table*}

\myparagraph{Baselines.} We compare to classical and recent approaches applicable to \task, divided between data-free and data-driven methods. 
For data-free baselines, we include \textbf{One-shot Magnitude Pruning} (\textbf{\textsc{omp}}, \cite{han2015deep}), shown to outperform several state-of-the-art algorithms after training \cite{frankle2020pruning} and comparable to IMP for pretrained transformers~\cite{jaiswal2023emergence}.  
We then include \textbf{\textsc{lamp}} \cite{lee2021layeradaptive}, which extends \textsc{omp} by unifying layer-wise calibration and global pruning. 
For data-driven methods, we choose two PaIs and two post-training pruning methods. 
Among PaIs, we select \textbf{\textsc{snip}} \cite{lee2018}, which retains connections based on their estimated impact on the loss function, and 
\textbf{\textsc{itersnip}} \cite{Jorge2020ProgressiveST} which gradually prunes by applying \textsc{snip} iteratively.
For post-training pruning methods, we test \textbf{\textsc{chita}}, a recent state-of-the-art algorithm that relies on a low-rank decomposition of the Hessian matrix of the loss function \cite{benbaki2023fast}, together with its iterative variant \textbf{\textsc{chita++}} \cite{benbaki2023fast}.
Additional baselines can be found in Sec.~\ref{sec:added-exps}.

\myparagraph{Experimental setup.}
We test all pruning methods at $63\%$ and $75\%$ global sparsity for all tasks, choosing them to exceed trivial sparsities (\ie, $\leq50\%$~\cite{gordon2020compressing}) and test in the neighborhood of the \emph{essential sparsity} (\ie, the limit after which the performance drop always overcomes the sparsity gain \cite{jaiswal2023emergence}, $\sim 70\%$).
We study the extreme $90\%$ sparsity in the next section. 
Data-dependent methods use the same set of calibration data on a \emph{per-run} basis.
To ensure no task-specific data is used for pruning, we construct $\mathcal{D}_g$ from CC3M \cite{sharma2018conceptual} and VisualGenome \cite{krishna2017visual}, and discard image-question pairs from the latter since they would collide with a portion of the data used for VQA finetuning.
We sample $B=3000$ batches from both datasets with a batch size of $b=32$, totaling around $\sim 5\%$ of the standard 4M pretraining set for VLMs ($\mathcal{D}_p$) \cite{ordonez2011im2text}.
In this way, a critical requirement of \task is satisfied, \ie, $|\mathcal{D}_g| << |\mathcal{D}_p|$.

For first- and second-order methods, we use general-purpose pretraining losses, as defined in the original papers, excluding objectives that require fine-grained annotations not always available (\ie, visual grounding in XVLM). 
Note that these methods already have some form of task prior and are, thus, expected to outperform the others: for both XVLM and BLIP, the pretraining loss contains either all or a subset of the finetuning losses.
For all downstream tasks, we fine-tune the pruned models with the same setup of the original papers and average the results over 3 runs with different seeds.  
We report additional details in the Appendix.

\subsection{Image-Text Retrieval (ITR)} \label{sec:itr}
\myparagraph{Setup.} 
We evaluate all methods and architectures for Image-Text Retrieval (ITR) on MSCOCO~\cite{karpathy2015deep}, analyzing both Text-to-Image retrieval, where the model should pick a matching image from a pool of target ones given a textual query, and the specular task of Image-to-Text retrieval.

\myparagraph{Results.} 
Results are summarized in Tab. \ref{tab:itr}, where we report the established Image-Recall (IR) and Text-Recall (TR) metrics at different levels. 
\method outperforms prior methods at all image-grounded and text-grounded metrics, at all sparsity levels, and for all models.
With XVLM, the gap between \method and the second best increases with the sparsity: at $63\%$ sparsity, \method outperforms \textsc{chita++} by $+0.76\%$ and $+0.90\%$ on TR@1 and IR@1 respectively, up to $+3.54\%$ and $+2.62\%$ when the overall sparsity increases to $75\%$.
{When pruning BLIP, \method remains the best performer while \textsc{omp} becomes the second best method, outperforming \textsc{chita++} at the lower sparsity.
In general, performance drops with BLIP are much larger than those of XVLM, with 3 out of 7 methods (denoted by \textsuperscript{$\ast$}) performing comparably to the random baseline.}
This hints at the following: \emph{given a fixed task, different VLMs expose different "prunabilities"}. 
Beyond the comparison between \method and the state-of-the-art, these experiments unveil another important finding: \emph{general purpose knowledge is retained by most methods}.
This observation is signaled by the TR@5 and IR@5 metrics, where the performance drop is contained regardless of the algorithm. %

\subsection{Image Captioning (IC)} \label{sec:ic}
\myparagraph{Setup.} Image Captioning (IC) is the task of generating a text given a source image (and a prompt in VLMs). %
We employ COCO Captions to evaluate the pruned models for this task \cite{chen2015microsoft}.
Following \cite{li2022blip, zeng2022multi}, during fine-tuning both BLIP and XVLM optimize a language-modeling loss %
and use the prompt \texttt{"a picture of"}. %
For quantitative evaluation, we use the standard BLEU@4 \cite{papineni2002bleu} and CIDEr \cite{vedantam2015cider} scores.
We report METEOR \cite{banerjee2005meteor} and SPICE \cite{anderson2016spice} in the Appendix. 

\myparagraph{Results.} The outcome of this experiment is reported in Tab. \ref{tab:vqa-cap}.
Notably, a pattern observed in ITR is confirmed also for IC: with XVLM, the gap between \method and the second best performing algorithm increases with the sparsity ($+0.50\%$ BLEU@4 and $+2.46$ CIDEr at $63\%$ becoming $+1.06\%$ and $+4.47$ at $75\%$ sparsity).
When pruning BLIP, \method and \textsc{chita++} perform equally well, with an edge for \method at $63\%$ sparsity and the opposite at $75\%$. 
Importantly, we make a key observation: for both ITR and IC, \emph{there is no fixed second best}. 
Different algorithms overcome each other when changing either the target model or sparsity.
This further underlines 
the importance of designing \emph{ad-hoc} methods for \task.

\subsection{Visual Question Answering (VQA)} \label{sec:vqa}
\myparagraph{Setup.}
VQA requires the model to answer a question by analyzing the content of an associated image.
The training set for the task collates the train and validation splits from the VQA2.0 dataset \cite{goyal2017making}, and the image-question pairs from Visual Genome \cite{krishna2017visual}, as customary in the field.
Given that generative capabilities are implicitly embodied by IC, with VQA we focus on the \emph{analytical capabilities} of the pruned models. 
Thus, we evaluate closed-set VQA, and let pruned models choose from a predefined set of 3129 answers \cite{li2021align}.

\begin{table*}[!ht]
 \centering
 \adjustbox{width=2\columnwidth}{
 {\scriptsize
 \begin{tabularx}{2\columnwidth}{
 >{\raggedright\arraybackslash}p{1.2cm}
 >{\centering\arraybackslash}p{1.2cm}
 >{\centering\arraybackslash}X
 >{\centering\arraybackslash}X
 >{\centering\arraybackslash}X
 >{\centering\arraybackslash}X
 >{\centering\arraybackslash}X
 >{\centering\arraybackslash}X
}
 \toprule
 \multirow{4}{*}{Method} & \multirow{4}{*}{Sparsity} & \multicolumn{3}{c}{{BLIP\textsubscript{BASE}}} & \multicolumn{3}{c}{{XVLM\textsubscript{CLIP}}} \\
 
 \cmidrule(lr){3-5}
 \cmidrule(lr){6-8}
 & & \multicolumn{1}{c}{VQA} & \multicolumn{2}{c}{Image Captioning} & \multicolumn{1}{c}{VQA} & \multicolumn{2}{c}{Image Captioning} \\
 \cmidrule(lr){3-3}
 \cmidrule(lr){4-5}
 \cmidrule(lr){6-6}
 \cmidrule(lr){7-8}

 & & \emph{test-dev} & BLEU@4 & CIDEr & \emph{test-dev} & BLEU@4 & CIDEr \\
 \midrule
 
 \textsc{dense} & $0\%$ & \cellcolor{lightgreen}76.31 &\cellcolor{lightgreen} 39.10 &\cellcolor{lightgreen} 131.14 & \cellcolor{lightorange} 76.92 &\cellcolor{lightorange} 38.99 &\cellcolor{lightorange} 130.43 \\ 

 \textsc{random} & \multirow{8}{*}{$63\%$} & 54.22 &15.88 &38.54 & 61.48 & 23.73 & 69.00 \\ 
 \textsc{snip} & &71.66$\pm$0.07 & 36.52$\pm$0.01 & 117.46$\pm$0.26 & 72.62$\pm$1.45 & 37.30$\pm$0.39 & 122.98$\pm$0.82 \\ 
 \textsc{itersnip} & & 63.26$\pm$0.01 & 26.57$\pm$0.10 & 75.80$\pm$0.26 & 72.92$\pm$0.14 & 36.52$\pm$0.16 & 120.12$\pm$0.52 \\ 
 \textsc{omp} & & \underline{73.59$\pm$0.04} & 37.36$\pm$0.07 & 124.86$\pm$0.27& 75.16$\pm$0.94 & \underline{37.97$\pm$0.09} & \underline{126.13$\pm$0.12} \\ 
 \textsc{lamp} & & 72.16$\pm$0.01 & 35.96$\pm$0.02 & 119.23$\pm$0.03 & 75.37$\pm$0.03 & 37.83$\pm$0.18 & 125.74$\pm$0.29 \\
 \textsc{chita} & &73.34$\pm$0.06 & 37.26$\pm$0.06 & 124.76$\pm$0.18 & \underline{75.82$\pm$0.04} & 37.82$\pm$0.20 & 125.75$\pm$0.57 \\
 \textsc{chita++} & & 73.52$\pm$0.03 & \underline{37.52$\pm$0.05} & \textbf{125.54$\pm$0.10} & 75.82$\pm$0.08 & 37.71$\pm$0.11 & 125.44$\pm$0.54 \\
 \rowcolor{multiflowcolor}
 \method & & \textbf{73.74$\pm$0.08} & \textbf{37.74$\pm$0.17} & \underline{125.40$\pm$0.29} & \textbf{76.02$\pm$0.03} & \textbf{38.47$\pm$0.01} & \textbf{128.59$\pm$0.13} \\
 
 \midrule

 \textsc{random} & \multirow{8}{*}{$75\%$} & 51.78 & 13.37 & 30.16 & 58.68 & 20.51 & 56.35 \\
 \textsc{snip} & & 67.07$\pm$0.13 & 33.5$\pm$0.21 & 101.29$\pm$0.62 & 69.77$\pm$0.03 & 34.29$\pm$0.27 & 110.22$\pm$1.41 \\ 
 \textsc{itersnip} & & 52.28$\pm$0.10 & 14.35$\pm$0.23 & 29.88$\pm$0.28 & 67.11$\pm$0.08 & 31.15$\pm$0.32 & 98.24$\pm$1.74 \\ 
 \textsc{omp} & & 69.41$\pm$0.07 & 34.92$\pm$0.05 & 113.77$\pm$0.24 & 73.48$\pm$0.07 & 35.74$\pm$0.09 & 118.37$\pm$0.29 \\
 \textsc{lamp} & & 63.33$\pm$0.78 & 28.2$\pm$0.08 & 88.25$\pm$0.12 & 73.22$\pm$0.02 & \underline{36.34$\pm$0.01} & \underline{120.13$\pm$0.13} \\
 \textsc{chita} & & 69.08$\pm$0.02 & 35.13$\pm$0.07 & 114.11$\pm$0.14 & 73.44$\pm$0.02 & 35.98$\pm$0.39 & 119.12$\pm$1.05 \\
 \textsc{chita++} & & \textbf{70.13$\pm$0.05} & \textbf{35.77$\pm$0.07} & \textbf{116.97$\pm$0.14} & \underline{74.00$\pm$0.01} & 36.07$\pm$0.04 & 119.48$\pm$0.16 \\
 \rowcolor{multiflowcolor}
 \method & & \underline{70.09$\pm$0.03} & \underline{35.73$\pm$0.10} & \underline{116.31$\pm$0.13}& \textbf{74.81$\pm$0.06} & \textbf{37.40$\pm$0.06} & \textbf{124.60$\pm$0.08} \\
 \bottomrule
 \end{tabularx}
 }
 }
 \caption{Results on VQA2.0 and COCO Captions at $63\%$ and $75\%$ sparsity. The dense \colorbox{lightgreen}{\textsc{blip}} and \colorbox{lightorange}{\textsc{xvlm}} upper bounds are reported on top. For further context, we include the random baseline as the lower bound. The \textbf{best performer} is bold; the \underline{second best} is underlined.}
 \label{tab:vqa-cap}
\end{table*}

\myparagraph{Results.} We evaluate all methods against the official evaluation website and provide results in Tab. \ref{tab:vqa-cap}.
VQA evaluation confirms the patterns: \method is the best performer in 3 out of 4 instances.
Within these 4 instances, we emphasize the further absence of a fixed second-best method: (1) \textsc{omp} is the second best with BLIP at $63\%$ sparsity; (2) at the same level, \textsc{chita} and \textsc{chita++} perform on par as the second best with XVLM; (3) at $75\%$ \method lies less than one standard deviation below the best performer \textsc{chita++} 
and (4) the latter underperforms the proposed method by $-0.81\%$ with XVLM.
These results disclose a complementary finding to Sec. \ref{sec:itr}: not only VLMs, but also \emph{vision-language tasks expose different prunabilities}.
In the next section, we further investigate both phenomena.

\section{Additional Analyses} \label{sec:abl}

\subsection{Extreme sparsity and different prunability}
\begin{figure*}[!t]
\setlength{\abovecaptionskip}{2pt}
\setlength{\belowcaptionskip}{0pt}
\centering
\includegraphics[width=2\columnwidth]{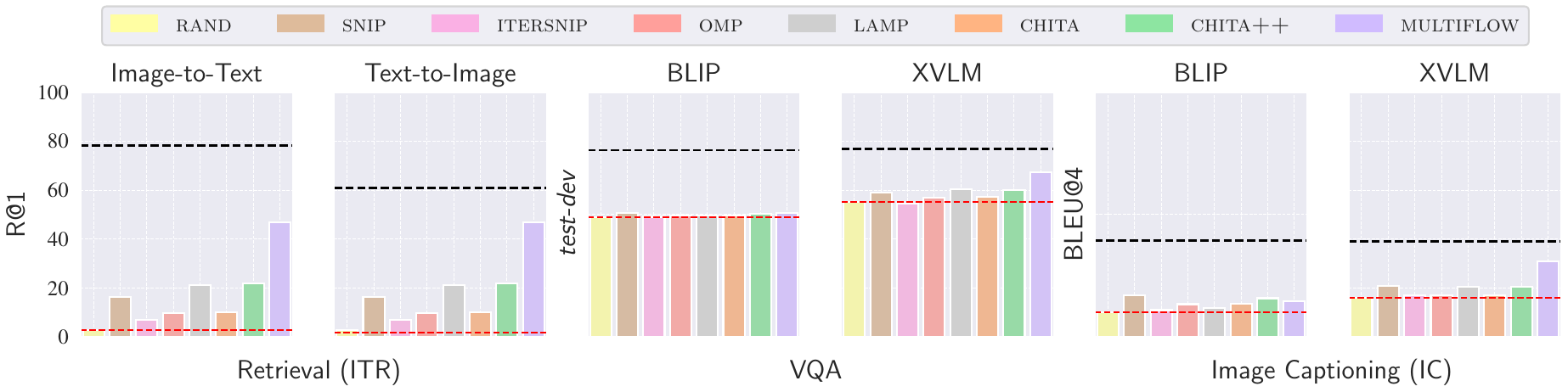}
\caption{Experiments at $90\%$ sparsity. ITR with XVLM (left) - VQA with both BLIP and XVLM (center) - IC with both BLIP and XVLM (right). The \textbf{{\color{lightred}\dashuline{random}}} and \textbf{\dashuline{dense}} baselines are also reported. All experiments follow the same configuration as those of Tabs. \ref{tab:itr} and \ref{tab:vqa-cap}.}
\label{fig:all_90}
\end{figure*}

In Sec. \ref{sec:experiments}, we report results at $63\%$ and $75\%$ sparsity, revealing that different VLMs and vision-language tasks exhibit inherently different prunabilities.
Here, we verify if these observations remain valid at $90\%$ global sparsity, an extreme compression level where performance reliable for real-world applications has not yet been reached in VLM pruning.
All of these experiments are depicted in Fig.~\ref{fig:all_90}.

\myparagraph{VLMs are not equally prunable}.
Results in Fig. \ref{fig:all_90} convey a strong message: even if identically pruned and similar in total parameter counts, every pruned BLIP model underperforms the corresponding pruned XVLM although starting from a generally better or comparable dense performance.
No method can produce meaningful results when pruning BLIP at $90\%$ sparsity regardless of the task, with all methods failing in ITR (\ie, $R@1 \leq 1\%$) and being almost on par with the random baseline in VQA and IC.
We hypothesize that parameters integrating different modalities are a key aspect to preserve, and that the explicit disentanglement among vision, text, and fusion modalities within XVLM makes it less sensitive to parameter removal. 
Here, \method outperforms \textsc{chita++}, the average second best, by +6.93, +10.14, and +24.88 in VQA-acc, BLEU@4, and TR@1, respectively, while also being $41\times$ faster on average, (see Tab.~\ref{tab:runtimes} in Appendix \ref{sec:additional-analyses}). 
We believe this is a great leap forward in pruning VLMs to extreme sparsities.

\myparagraph{Algorithms rank differently across settings.}
The observation that \emph{no fixed second best} is present at $63\%$ and $75\%$ sparsities also remains valid at $90\%$.
For example, \textsc{lamp} outperforms \textsc{snip} in ITR, while the opposite happens in IC.
This hints at the fact that pruning algorithms preserve different types of encoded knowledge and further underlines the effectiveness of the task-agnostic design of \method which steadily maintains good performance across tasks, outperforming all competitors in all tasks for XVLM.

\myparagraph{VL tasks can be ranked by difficulty.} 
The \emph{prune-then-transfer} paradigm leads to extreme differences in how performance drops according to the downstream task.
With XVLM, the gaps between the dense upper bound and the best performer \method are $\downarrow\text{TR@1}\sim40\%$ and $\downarrow\text{IR@1}\sim43\%$ for ITR, but jump to only $\downarrow\text{\emph{test-dev}-acc}\sim12\%$ and $\downarrow BLEU\text{@}4\sim20\%$ for VQA and IC.
While this may highlight different sensitivities among metrics, it also enables ranking the tasks as VQA $<$ IC $<$ ITR in terms of their difficulty, hinting that a good algorithm for \task shall emphasize image-text alignment. 

\subsection{Ablations and sanity checks on \method}
In this section we perform two ablation studies of \method, testing the performance of the inverted mask, and ablating the impact of the imposed distribution. 

\myparagraph{Inversion.} Inspired by \cite{frankle2020pruning}, we check if the saliency function at the core of a score-based pruning algorithm effectively extracts the most important weights, by inverting the mask and maintaining the parameters with the lowest scores.
Tab \ref{tab:ablations} shows that the score in \method successfully does so: inverting the pruning mask even extracts subnetworks that perform far \emph{worse} than the random baseline. 

\begin{table}
 \centering
 {\scriptsize
  \begin{tabular}{cccc}
   \toprule
   Ablation & Sparsity & Text R@1 & Image R@1 \\ 
   \midrule
    \rowcolor{lightgray} \textsc{random} &\multirow{5}{*}{$63\%$}& 21.48 & 14.89 \\

   \rowcolor{lightgray}  \textsc{multiflow} w/ Inversion & \ & 7.15 & 5.16 \\ 
  \method w/o distribution & & 0.01$\pm$0.01 & 0.02$\pm$0.0 \\
     \method w/o \emph{multimodality} & & 76.66$\pm$0.09 & 59.96$\pm$0.20 \\ 
    \rowcolor{multiflowcolor}\textsc{multiflow} & & 77.35 & 60.21 \\ 
   \midrule
     \rowcolor{lightgray}\textsc{random}&\multirow{5}{*}{$75\%$}  & 12.18 & 8.63 \\

    \rowcolor{lightgray} \textsc{multiflow} w/ Inversion & & 1.91 & 0.88 \\
     \method w/o distribution & & 0.02$\pm$0.0 & 0.01$\pm$0.01 \\
     \method w/o \emph{multimodality} & & 72.50$\pm$0.05 & 55.75$\pm$0.12 \\ 
    \rowcolor{multiflowcolor}\textsc{multiflow} & & 73.87 & 56.94 \\
   \midrule
   
    \rowcolor{lightgray} \textsc{random} & \multirow{5}{*}{$90\%$}& 2.92 & 1.93 \\
    \rowcolor{lightgray}  \textsc{multiflow} w/ Inversion &  & 0.43 & 0.20 \\
      \method w/o distribution &  & 0.02$\pm$0.0 & 0.03$\pm$0.01 \\
     \method w/o \emph{multimodality}& & 35.55$\pm$0.26 & 25.00$\pm$0.06 \\ 
   \rowcolor{multiflowcolor}\textsc{multiflow} & & 46.87 & 34.77 \\
   \bottomrule
  \end{tabular}
}
 \caption{Ablation study on ITR and MSCOCO (\textsc{XVLM}). Both the \colorbox{lightgray}{inversion} and the \whitebox{imposed distribution} studies are reported.}
 \label{tab:ablations}
 \vspace{-10pt}
\end{table}

\begin{figure}
    \setlength{\abovecaptionskip}{0pt}
    \setlength{\belowcaptionskip}{0pt}
    \centering
    \includegraphics[width=1\columnwidth]{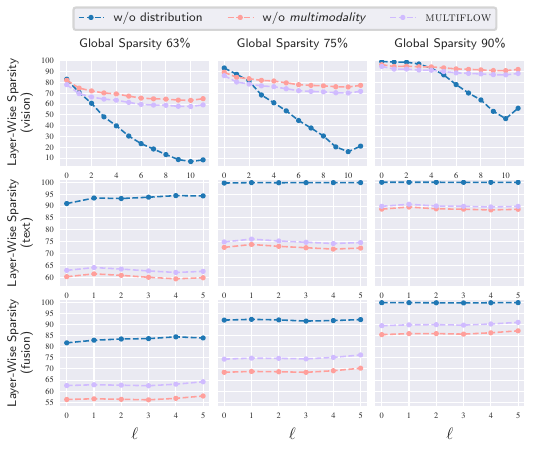}  
    \caption{Comparison of the sparsities obtained at each layer $\ell$ of each modality by (i) pruning with the $\mathtt{topk}$ global scores of \method (denoted by \emph{w/o distribution}), (ii)  \textsc{omp} (\emph{w/o multimodality}) and (iii) \method. The figure displays XVLM.}
    \label{fig:distributions}
    \vspace{-15pt}
\end{figure}

\myparagraph{Imposed distribution.} Here we test the impact of the imposed multimodal prior described in Sec.~\ref{sec:mmp} on \method, replacing it with two variants (i) not imposing a prior distribution on the layer-wise pruning ratios (\ie, the mask is computed taking the top-k global scores), (ii) determining the layer-wise pruning ratios according to the weight magnitude.
These distributions are displayed in Fig.~\ref{fig:distributions} for different parts of the model (\ie, vision, text, fusion), and with corresponding performance on ITR in Tab.~\ref{tab:ablations}. 
As we can see from Fig.~\ref{fig:distributions}, without imposing any prior distribution (blue line), the model would either heavily prune early layers (\eg, vision) or full modalities (\eg, text), resulting in performance even worse than random due to model collapse (Tab.~\ref{tab:ablations}). 
The magnitude-based distribution (\ie, w/o multimodality) already recovers this effect, with a more distributed pruning across layers and modalities. 
However, including multimodal priors leads to the best results, especially at 90\% sparsity (\eg, +11\% on TR@1, +9\% on IR@1). 
This experiment discloses two final findings: (i) considering potential biases in activation patterns among layers and modalities is fundamental for VLM pruning and (ii) in high-sparsity regimes, a small shift in the layer-wise distribution can correspond to a large performance gap.

\section{Conclusions}
In this work, we formalized and addressed \emph{\taskFull} (\task).
We proposed \methodFull (\method), an approach that preserves the information flow within the original VLM by exploiting a (multimodal) prior on the weight magnitude for layer-wise pruning, and by incorporating the saliencies of input/output nodes into the scoring criterion for network parameters.
We also benchmarked 8 pruning methods for \task, using two VLMs, three vision-language tasks, and different pruning ratios, showing that \method outperforms existing state-of-the-art methods in the vast majority of the cases.
While our results highlight a large gap between the dense model and its pruned counterpart, this work is a step toward finding task-agnostic pruned VLMs.
\label{sec:concl}

{\small
\myparagraph{Acknowledgements.} 
We acknowledge the CINECA award under the ISCRA initiative for the availability of high-performance computing resources and support.
E.R. and M.M. are supported by the MUR PNRR project FAIR - Future AI Research (PE00000013), funded by NextGeneration EU. E.R. is also supported by the EU projects AI4TRUST (No.101070190) and ELIAS (No.01120237). M.F. is supported by the PRIN project LEGO-AI (Prot.2020TA3K9N) and the PAT project ``AI@TN".
This work was also partially sponsored by a Cisco Research Grant.}

\clearpage

{
 \small
 \bibliographystyle{ieeenat_fullname}
 \bibliography{main}
}

\clearpage
\appendix
\setcounter{page}{1}
\maketitlesupplementary

\newcommand{\tsep}{\texttt{|}\xspace}
\newcommand{\omp}{\textsc{omp}\xspace}
\newcommand{\tamt}{\textsc{tamt}\xspace}
\newcommand{\lamp}{\textsc{lamp}\xspace}
\newcommand{\imp}{\textsc{imp}\xspace}
\newcommand{\lTwo}{$\ell_2$\textsc{-norm}\xspace}
\newcommand{\mt}{\textsc{mt}\xspace}

In this Supplementary Material we complement the content of the main paper.
The document is structured as follows: Sec.~\ref{sec:added-exps} reports comparisons with more baselines and an additional unimodal experiment; Sec. \ref{sec:impl-details} contains all implementation details of Sec. \ref{sec:experiments} of the main paper; in Sec. \ref{sec:additional-analyses} we analyze the pruning efficiency of each algorithm in terms of runtime and extend the discussions of Sec. \ref{sec:abl} of the main paper; in Sec. \ref{sec:limitations} we discuss the limitations as well as the broader impacts of the proposed setting and method. 

\section{Additional Experiments}
\label{sec:added-exps}

{

\subsection{Additional baselines}

\myparagraph{Mask Training.}
Here we compare \method with a procedure requiring much more compute: Task-Agnostic Mask Training (\textsc{tamt}) \cite{liu2022learning}. 
In a nutshell, \textsc{tamt} works by optimizing the pruning masks as trainable parameters during the pretraining phase with straight-through gradient estimation.
In its default setup, \textsc{tamt} would require the entire pretraining dataset $D_p$ and many optimization epochs, thus going against the principles of the lightweight pruning strategy proposed in the main paper. 
To meet the constraints of TA-VLP and to ensure a fair comparison, for this experiment we fix the computational budget of \textsc{tamt} to that of gradient-based pruning algorithms in our competitor suite, meaning that we effectively train the pruning masks on $\sim5\%$ of the 4M pretraining dataset for 1 epoch. 

\myparagraph{Layer-wise $\ell_2$ normalization.}
Additionally, we examine a different way to unify global magnitude pruning and layer-wise calibration other than \lamp, \ie, normalizing the score tensor of each layer by its Frobenius norm before global pruning (which is equivalent to layer-wise normalization by the $\ell_2$ norm of the flattened score tensor).
The scoring criterion before normalization is the weight magnitude, as in \lamp.
This procedure was shown effective in the context of CNNs \cite{molchanov2017pruning}, so we aim to shed light on its performance with large-scale transformers.

\myparagraph{Results} are reported in Tab.~\ref{tab:addedBaselines} for XVLM at all compression levels, from which we can observe that \tamt is an effective procedure for moderated sparsity levels, but becomes comparable to even a data-free procedure such as \lamp at extreme compression rates, thus hardly justifying the additional compute it requires. 
A final observation is that a simple layer-wise $\ell_2$ normalization is not sufficient for accurate pruning of large-scale VLMs.

A comparison between \method and \tamt for pruning BLIP is in Tab.~\ref{tab:addedBaselinesBLIP}. 
Consistently with the results of the main paper, also \tamt fails in pruning BLIP at great sparsities, so the table only reports $63\%$ and $75\%$ results.
``ERR" indicates that the pruned model did not converge to meaningful results, with an average R@1 $\sim 0\%$.
In general, the insights of this experiment do not diverge from the observations of the main paper, \ie: \ding{172} pruning BLIP is more challenging than pruning XVLM also for \tamt; \ding{173} \method is more efficient and outperforms \tamt across all tasks; \ding{174} the gap between \method and \tamt increases with the sparsity, regardless of the downstream task.

\begin{table*}
    \centering
    {
    \begin{tabularx}{2\columnwidth}{
    >{\hsize=.3\hsize\raggedright\arraybackslash}X
    >{\hsize=.425\hsize\centering\arraybackslash}X
    >{\hsize=.425\hsize\centering\arraybackslash}X
    >{\hsize=.425\hsize\centering\arraybackslash}X
    >{\hsize=.425\hsize\centering\arraybackslash}X
    }
    
    \toprule
    \multirow{2.5}{*}{Method} & \multicolumn{2}{c}{Image-Text Retrieval} & \multicolumn{2}{c}{Image Captioning} \\

    \cmidrule(lr){2-3}
    \cmidrule(lr){4-5}
     & Text R@1 & Image R@1 & BLEU@4 & CIDEr \\
    \midrule

    \rowcolor{lightorange}
    \textsc{xvlm\textsubscript{clip}} & 78.2 & 60.9 & 39.0 & 130.4 \\
    
    \lTwo & 74.5 \tsep 67.9 \tsep 14.8 & 57.6 \tsep 51.1 \tsep 10.8 & \underline{37.9} \tsep 35.7 \tsep 18.6 & 125.7 \tsep 117.6 \tsep 48.0 \\
    \lamp & 75.3 \tsep 69.4 \tsep 21.4 & 58.4 \tsep 53.2 \tsep 14.6 & 37.8 \tsep 36.3 \tsep \underline{20.7} & 125.7 \tsep 120.1 \tsep \underline{56.9} \\
    
    \textsc{\tamt} & \underline{76.8} \tsep \underline{72.2} \tsep \underline{21.6} & \underline{59.9} \tsep \underline{55.6} \tsep 14.6 & 37.8 \tsep \underline{36.7} \tsep 20.2 & \underline{126.2} \tsep \underline{122.1} \tsep 54.1 \\

    \rowcolor{multiflowcolor}
    \method & \textbf{77.4} \tsep \textbf{73.9} \tsep \textbf{46.9} & \textbf{60.2} \tsep \textbf{56.9} \tsep \textbf{34.8} & \textbf{38.5} \tsep \textbf{37.4} \tsep \textbf{30.8} & \textbf{128.6} \tsep \textbf{124.6} \tsep \textbf{94.3} \\

    \bottomrule
    \end{tabularx}
    }
    \caption{Additional baselines on \textsc{xvlm}, formatted as $63$\texttt{|}$75$\texttt{|}$90$ sparsities [$\%$]. We also report \lamp to facilitate comparing with $\ell_2$-\textsc{norm}.}
    \label{tab:addedBaselines}
\end{table*}

\begin{table*}
    \centering
    {
    \begin{tabularx}{2\columnwidth}{
    >{\hsize=.3\hsize\raggedright\arraybackslash}X
    >{\hsize=.425\hsize\centering\arraybackslash}X
    >{\hsize=.425\hsize\centering\arraybackslash}X
    >{\hsize=.425\hsize\centering\arraybackslash}X
    >{\hsize=.425\hsize\centering\arraybackslash}X
    }
    
    \toprule
    \multirow{2.5}{*}{Method} & \multicolumn{2}{c}{Image-Text Retrieval} & \multicolumn{2}{c}{Image Captioning} \\

    \cmidrule(lr){2-3}
    \cmidrule(lr){4-5}
     & Text R@1 & Image R@1 & BLEU@4 & CIDEr \\
    \midrule

    \rowcolor{lightgreen}
    \textsc{blip\textsubscript{base }} & 80.7 & 95.1 & 39.1 & 131.1 \\
    
    \textsc{\tamt} & 75.2 \tsep ERR & 57.9 \tsep ERR & 37.3 \tsep 34.1 & 124.0 \tsep 111.5 \\

    \rowcolor{multiflowcolor}
    \method & \textbf{76.3} \tsep \textbf{65.7} & \textbf{59.0} \tsep \textbf{49.9} & \textbf{37.7} \tsep \textbf{35.7} & \textbf{125.4} \tsep \textbf{116.3} \\

    \bottomrule
    \end{tabularx}
    }
    \caption{Comparison between \tamt and \method on \textsc{blip}, formatted as $63$\texttt{|}$75$ sparsities [$\%$].}
    \label{tab:addedBaselinesBLIP}
\end{table*}

\subsection{Pruning Vision-only models}
\label{sec:dino-exps}
While a core component of \method is the multimodality-aware injection of layer-wise pruning ratios, the method may also be applied to unimodal models with slight changes.
Specifically, since no modalities to disambiguate are present in these scenarios, then the pruning ratios can be directly injected from global magnitude pruning.
For any layer, this entails retaining as many parameters as one would get by applying global magnitude pruning, according to the scoring function of \method. 

\myparagraph{Setup.} We test this straightforward variation on a Vision Transformer \cite{dosovitskiy2020image} pretrained with the DINO method \cite{caron2021emerging}, by comparing it to \textsc{omp} and \textsc{tamt}.
Once the pruned models are obtained, we transfer them to 3 downstream datasets for image classification, namely CIFAR10, CIFAR100, and Flowers102 \cite{nilsback2008automated, krizhevsky2009learning}.
Following Sec.~\ref{sec:experiments} of the main paper, \method and \textsc{tamt} use $\sim5\%$ of the model's pretraining set for pruning (\ie, 64k images from ImageNet-1k \cite{deng2009imagenet}).
We also test \method under extreme data scarcity, using only 128 images for pruning, and denote the results with \textsc{mf\textsubscript{128}}.
All pruned models are trained on 224$\times$224 images for 10 epochs with a batch size of $b=32$, and test performance after training is reported.
We use AdamW with $\beta = (0.9, 0.999)$, and cosine learning rate decay with a warmup to $1\times10^{-5}$ during the first $10\%$ of the training steps.

\myparagraph{Results} are in Tab.~\ref{tab:dino} for all sparsity levels. 
Notably, \method outperforms \textsc{tamt} even with 500x less data and, on a general note, it  
outperforms the selected baselines for this additional experiment.
This result suggests that the scoring function proposed in \method remains sound regardless of the target model to prune, and we hope this additional insight may be of use for future research on network pruning.

We also observe that \method is robust to data scarcity: pruning with only 128 images is comparable to using much more data in most cases. Enlarging the size of $D_g$ tends to be more useful when the task is ``difficult", \ie, it is more useful on CIFAR100 than it is on CIFAR10, or when the sparsity is greater. }

\begin{table*}
    \centering
    \small
        
    \begin{tabularx}{2\columnwidth}{
    >{\hsize=.4\hsize\raggedright\arraybackslash}X
    >{\hsize=.5\hsize\centering\arraybackslash}X
    >{\hsize=.5\hsize\centering\arraybackslash}X
    >{\hsize=.6\hsize\centering\arraybackslash}X
    }
   
    \toprule
    Method & Cifar10 & Cifar100 & Flowers102 \\
    \midrule

    \rowcolor{lightgray}
    \textsc{dino\textsubscript{B/16}} & 99.1 & 91.7 & 98.8 \\
    \textsc{\textsc{omp}} & 98.0 \tsep 94.7 \tsep 58.1 & 83.5 \tsep 66.7 \tsep 23.9 & 92.9 \tsep 78.0 \tsep 22.3 \\
    \textsc{\textsc{tamt}} & 97.8 \tsep 95.8 \tsep 57.9 & 77.2 \tsep 69.2 \tsep 22.8 & 92.3 \tsep 84.4 \tsep 21.8 \\

    \rowcolor{multiflowcolor}
    \textsc{multiflow}\textsubscript{$|D_g|=128$} & \textbf{98.1} \tsep \textbf{96.4} \tsep \underline{66.4} & \underline{84.9} \tsep \underline{74.0} \tsep \underline{28.7} & \textbf{96.3} \tsep \underline{87.1} \tsep \underline{30.4} \\

    \rowcolor{multiflowcolor}
    \method & \textbf{98.1} \tsep \underline{96.3} \tsep \textbf{67.4} & \textbf{85.6} \tsep \textbf{74.6} \tsep \textbf{29.2} & \underline{96.2} \tsep \textbf{87.5} \tsep \textbf{30.5} \\

    \bottomrule
    \end{tabularx}
\caption{Image Classification with pruned \textsc{dino} models. Formatted as $63$\texttt{|}$75$\texttt{|}$90$ sparsities [$\%$].}
\label{tab:dino}
    
\end{table*}

\section{Implementation Details}\label{sec:impl-details}

\myparagraph{Remark.} 
It is not straightforward to %
test existing VLM pruning methods \cite{shi2023upop,wang2022efficientvlm} in \task. First, $|\mathcal{D}_g| << |\mathcal{D}_p|$, while \cite{wang2022efficientvlm} entails pretraining a compact VLM on the entire $\mathcal{D}_p$ with knowledge distillation from the dense VLM.
Second, both \cite{shi2023upop} and \cite{wang2022efficientvlm} require task-specific knowledge since they prune using the actual target downstream datasets during training, while we design $\mathcal{D}_g$ to be explicitly task-agnostic and prune at pretrained initialization.
Finally, \cite{shi2023upop} and \cite{wang2022efficientvlm} target structured pruning, while our benchmarking on \task is focused on unstructured pruning, since the latter category generally leads to better performance (see, \eg, \cite{chen2021chasing}).
Thus, to avoid unfair comparisons, we acknowledge but do not compare with \cite{shi2023upop, wang2022efficientvlm}.

\begin{table}[!t]
 \centering
 \adjustbox{width=\columnwidth}{
 \begin{tabularx}{\columnwidth}{
 >{\raggedright\arraybackslash}p{1.75cm}
 >{\centering\arraybackslash}X
 >{\centering\arraybackslash}X
}

 \toprule
 \multirow{2.5}{*}{Method} & \multicolumn{2}{c}{Runtime [$s$]} \\

 \cmidrule(lr){2-3}
 & {BLIP} & {XVLM} \\
 \midrule

 \textsc{snip} & 2770.61 {\footnotesize ($\times2.82$)} & 1847.04 {\footnotesize ($\times2.67$)}\\
 \textsc{itersnip} & 2963.97 {\footnotesize ($\times3.02$)} & 1966.03 {\footnotesize ($\times2.85$)}\\
 \textsc{chita} & 8351.92 {\footnotesize ($\times8.51$)} & 3879.94 {\footnotesize ($\times5.62$)}\\
 \textsc{chita++} & 40538.03 {\footnotesize ($\times41.30$)} & 28368.73 {\footnotesize ($\times41.08$)}\\
 \rowcolor{multiflowcolor}\method & 981.45 & 690.64 \\

 \bottomrule
 
 \end{tabularx}
 }
 \caption{Mean runtime [$s$] of data-driven pruning algorithms, computed over the sparsity levels of Secs. \ref{sec:experiments} and \ref{sec:abl} of the main paper. In brackets the speedup of \method.}
 \vspace{-1.5em}
 \label{tab:runtimes}
\end{table}

\subsection{Experimental methodology}
In this section, we provide %
additional details on our pruning, training and evaluation methodologies. %

\myparagraph{Usage of calibration data.}
For one-shot, data-driven methods (\ie, \textsc{snip} and \method) we use the $B = 3000$ batches from $\mathcal{D}_g$.
For \textsc{chita}, execution is bound by hardware availability since it requires keeping an FP32 matrix of size $B \times |\Theta|$ into volatile memory. 
As a consequence, only for this algorithm, we use $B=500$ batches from both datasets, accounting for a total of $\sim 368\text{GB}$ and $\sim 453\text{GB}$ of RAM usage to store such a matrix for XVLM and BLIP, respectively\footnote{We experiment on compute nodes with $\sim$490GB RAM, and exceeding the limit of $B=500$ batches can lead to OOM errors with BLIP.}.
For all iterative methods (\ie, \textsc{itersnip} and \textsc{chita++}), we follow the experimental setup of \cite{Jorge2020ProgressiveST} and fix the computational budget of their one-shot counterparts.
Thus, we use $T = 60$ pruning iterations and $B_t = 50$ batches per iteration, such that $T \times B_t = 60 \times 50 = 3000 = B$. 
Both algorithms use an exponential sparsity schedule, as suggested by the corresponding authors \cite{Jorge2020ProgressiveST,benbaki2023fast}.

\myparagraph{Peculiarities when pruning.} 
BLIP \cite{li2022blip} is composed of (i) a vision-encoder and (ii) a mixture of encoder-decoder transformers (MED) for text encoding and text decoding. 
The text-encoder and decoder of the MED share \emph{all} the weights except for the self-attentions.
In our work, we follow this design choice and require that these properties are preserved by the pruned models.
As a consequence, we share the pruning masks according to the MED, \ie, shared weights also share the pruning masks. 
In contrast, XVLM \cite{zeng2022multi} comprises a vision-encoder, a text-encoder and a fusion encoder on top. 
No weights are tied, so no modules share their respective pruning masks.

\myparagraph{Extracting gradient information from $\mathcal{D}_g$.}
Among the algorithms benchmarked in Sec. \ref{sec:experiments}, \textsc{snip}, \textsc{itersnip}, \textsc{chita} and \textsc{chita++} require gradient information.
However, using a task-specific loss function to compute gradients is not possible in \task.
Therefore, for all the aforementioned algorithms, we use the general purpose objective encoded by the pretraining loss of each model ($\mathcal{L}_{p}$).
For BLIP, $\mathcal{L}_{p} = \mathcal{L}_{\text{ITC}} + \mathcal{L}_{\text{ITM}} + \mathcal{L}_{\text{LM}}$, where (i) $\mathcal{L}_{\text{ITC}}$ is the image-text contrastive loss introduced by \cite{radford2021learning}; (ii) $\mathcal{L}_{\text{ITM}}$ is the image-text matching loss, computed with binary cross-entropy between the outputs of the matching head and the groundtruth given by in-batch hard negative mining and (iii) $\mathcal{L}_{\text{LM}}$ is the language modeling loss, computed with the outputs of the image-grounded text decoder of the MED. 
For XVLM, we use $\mathcal{L}_p = \mathcal{L}_{\text{ITC}} + \mathcal{L}_{\text{ITM}} + \mathcal{L}_{\text{MLM}}$, where $\mathcal{L}_{\text{MLM}}$ is the masked language modeling loss on the outputs of the fusion encoder.
For additional details, please refer to the original papers \cite{li2022blip, zeng2022multi}.

\myparagraph{Tuning \textsc{chita} and \textsc{chita++} hyperparameters.}
Both \textsc{chita} and \textsc{chita++} rely on a ridge penalty, denoted by $\lambda$, to penalize deviations from dense weights. 
In line with the authors of these methods, we find both algorithms to be very sensitive to this hyperparameter. 
Hence, to ensure a fair comparison we run a grid search on $\lambda$ using ITR as a proxy task. 
Specifically, we train all pruned models at $63\%$ sparsity with $\lambda$ ranging from $10^{-5}$ to $10^3$ for one epoch, then pick the best value for the benchmark of the main paper.
We report full results of this tuning in Tab. \ref{tab:chita-hp-search}.

\myparagraph{Downstream task fine-tuning.} We train all the pruned models with 4 NVIDIA A100 GPUs and mixed-precision, then average the results over three different runs. %
All fine-tuning configurations are reported in Tab. \ref{tab:hps}, for each task.
We also re-execute all the dense baselines under the same setup, to ensure that hardware differences do not play any role when measuring the performance drops after pruning.

\section{Additional analyses}\label{sec:additional-analyses}

\subsection{Pruning efficiency.}
We report in Tab. \ref{tab:runtimes} the total runtime ($s$) of all data-driven methods (data-free methods can be executed in negligible time by most modern hardware). 
When pruning, we fix hardware requirements to only one NVIDIA A100 GPU for all algorithms, and use the FP32 data type.
As the table shows, \method is much faster than data-driven methods requiring gradient information and/or combinatorial optimization, being approximately $3\times$ faster than \textsc{snip} (second fastest) and $41\times$ faster than \textsc{chita++} (the slowest in our benchmark).

\subsection{On the importance of nodes and edges}
In Sec. \ref{sec:mfp} we introduce \method and its scoring criterion, which incorporates an explicit formulation for neuron-level importance as well as for the importance of individual connections.
While we demonstrate, in Sec. \ref{sec:abl}, that inverting the pruning mask in \method leads to even worse than random performance, thus verifying the soundness of our scoring function, here we assess the importance of each component of the score using ITR as a proxy task.
We report results in Tab. \ref{tab:score-abl-xvlm}.
We observe that using only the importance of edges (\ie, using the weight magnitude as the scoring criterion), performs better than using only node-level saliency. 
Despite this, we also observe that removing either component leads to large performance drops, confirming that both elements are beneficial for pruning.

\begin{table}
 \centering
 \begin{adjustbox}{width=\columnwidth} %
  \begin{tabular}{ccccc}
   \toprule
   Edge & Nodes & Sparsity & Text R@1 & Image R@1 \\ 
   
   \midrule
   \checkmark & \xmark & \multirow{3}{*}{$63\%$} & 76.97$\pm$0.32 & 59.51$\pm$0.12 \\
   \xmark & \checkmark & & 56.76$\pm$0.49 & 41.91$\pm$0.13 \\ 
   \rowcolor{multiflowcolor}
   \checkmark & \checkmark & & 77.35$\pm$0.51 & 60.21$\pm$0.16 \\ 
   
   \midrule
   \checkmark & \xmark & \multirow{3}{*}{$75\%$}& 72.34$\pm$0.37 & 55.73$\pm$0.08 \\
   \xmark & \checkmark & & 39.63$\pm$0.52 & 28.49$\pm$0.10 \\
   \rowcolor{multiflowcolor}
   \checkmark & \checkmark & & 73.87$\pm$0.13 & 56.94$\pm$0.10 \\
   
   \midrule
   \checkmark & \xmark & \multirow{3}{*}{$90\%$} & 14.37$\pm$0.53 & 10.59$\pm$0.17 \\
   \xmark & \checkmark & & 10.47$\pm$0.26 & 7.11$\pm$0.14 \\ 
   \rowcolor{multiflowcolor} 
   \checkmark & \checkmark & & 46.87$\pm$0.11 & 34.77$\pm$0.10 \\
   \bottomrule
  \end{tabular} 
 \end{adjustbox}
 \caption{Ablation of each component of the score (XVLM), computed considering only edges, only nodes, or both (as in the original {\textsc{multiflow}} algorithm).}
 \vspace{-1.5em}
 \label{tab:score-abl-xvlm}
\end{table}

\subsection{Image Captioning with METEOR and SPICE.}
For completeness, in Tab. \ref{tab:meteor-spice} we report the METEOR \cite{banerjee2005meteor} and SPICE \cite{anderson2016spice} scores of the experiments shown in Tab. \ref{tab:vqa-cap} of the main paper.
Notably, these metrics confirm the findings emerged with the BLEU \cite{papineni2002bleu} and the CIDEr \cite{vedantam2015cider} scores: \ie, 
\method outperforms all baselines for all models at $63\%$ sparsity, and becomes slightly worse than only \textsc{chita++} when pruning BLIP at $75\%$ sparsity.
Importantly, we observe that algorithms rank differently also with these metrics when the target VLM changes. 
For example, while \textsc{chita++} outperforms \textsc{lamp} when pruning BLIP, the latter performs better or comparably when pruning XVLM.

\begin{table*}[!ht]
 \centering
 \adjustbox{width=2\columnwidth}{
 {\footnotesize
 \begin{tabularx}{2\columnwidth}{
 >{\raggedright\arraybackslash}p{1.2cm}
 >{\centering\arraybackslash}p{1.2cm}
 >{\centering\arraybackslash}X
 >{\centering\arraybackslash}X
 >{\centering\arraybackslash}X
 >{\centering\arraybackslash}X
}
 \toprule
 \multirow{2.5}{*}{Method} & \multirow{2.5}{*}{Sparsity} & \multicolumn{2}{c}{{BLIP}} & \multicolumn{2}{c}{{XVLM}} \\

 \cmidrule(lr){3-4}
 \cmidrule(lr){5-6}
 & & METEOR & SPICE  & METEOR & SPICE \\
 \midrule
 
 \textsc{dense} & $0\%$ &\cellcolor{lightgreen} 30.63 &\cellcolor{lightgreen} 23.49 &\cellcolor{lightorange} 30.50 &\cellcolor{lightorange} 23.41 \\ 

 \textsc{random} & \multirow{8}{*}{$63\%$} & 16.35 & 9.04 & 21.35 & 14.14 \\ 
 \textsc{snip} & & 28.34$\pm$0.05 & 21.47$\pm$0.02 & 29.00$\pm$0.12 & 22.17$\pm$0.10 \\ 
 \textsc{itersnip} & & 22.42$\pm$0.04 & 15.17$\pm$0.02 & 28.58$\pm$0.06 & 21.76$\pm$0.05 \\ 
 \textsc{omp} & & 29.69$\pm$0.01& 22.70$\pm$0.01 & 29.76$\pm$0.04 & 22.74$\pm$0.04 \\ 
 \textsc{lamp} & & 28.82$\pm$0.02 & 21.97$\pm$0.01 & \underline{29.78$\pm$0.06} & \underline{22.79$\pm$0.01} \\
 \textsc{chita} & & 29.62$\pm$0.06 & 22.61$\pm$0.02 & 29.77$\pm$0.05 & 22.79$\pm$0.06\\
 \textsc{chita++} & & \underline{29.74$\pm$0.02} & \underline{22.74$\pm$0.02} & 29.65$\pm$0.07 & 22.71$\pm$0.08 \\
 \rowcolor{multiflowcolor}
 \method & & \textbf{29.75$\pm$0.04} & \textbf{22.82$\pm$0.07} & \textbf{29.96$\pm$0.03} & \textbf{23.07$\pm$0.05} \\
 
 \midrule

 \textsc{random} & \multirow{8}{*}{$75\%$} & 14.58 & 7.19 & 19.11 & 11.97 \\
 \textsc{snip} & & 25.98$\pm$0.11 & 18.91$\pm$0.07 & 27.02$\pm$0.22 & 20.24 $\pm$0.28 \\ 
 \textsc{itersnip} & & 14.19$\pm$0.06 & 6.92$\pm$0.05 & 25.51$\pm$0.25 & 18.52$\pm$0.27 \\ 
 \textsc{omp} & & 27.79$\pm$0.01 & 20.96$\pm$0.03 & 28.52$\pm$0.05 & 21.70$\pm$0.03 \\
 \textsc{lamp} & & 24.26$\pm$0.00 & 17.18$\pm$0.01 & \underline{28.73$\pm$0.01} & 21.76$\pm$0.02 \\
 \textsc{chita} & & 27.86$\pm$0.03 & 21.03$\pm$0.03 & 28.67$\pm$0.19 & 21.78$\pm$0.17 \\
 \textsc{chita++} & & \textbf{28.18$\pm$0.06} & \textbf{21.38$\pm$0.01} & 28.64$\pm$0.04 & \underline{21.79$\pm$0.06} \\
 \rowcolor{multiflowcolor}
 \method & & \underline{28.12$\pm$0.04} & \underline{21.27$\pm$0.02} &  \textbf{29.21$\pm$0.03} & \textbf{22.40$\pm$0.01} \\
 \bottomrule
 \end{tabularx}
 }
 }
 \caption{Image Captioning results with METEOR \cite{banerjee2005meteor} and SPICE \cite{anderson2016spice} scores. Naming and coloring follow Tab. \ref{tab:vqa-cap} of the main paper.}
 \label{tab:meteor-spice}
\end{table*}

\begin{table*}[!ht]
    \centering
    {\footnotesize
    \begin{tabularx}{2\columnwidth}{
    >{\centering\arraybackslash}p{1.2cm}
    >{\centering\arraybackslash}p{1cm}
    >{\centering\arraybackslash}X
    >{\centering\arraybackslash}X
    >{\centering\arraybackslash}X|
    >{\centering\arraybackslash}X
    >{\centering\arraybackslash}X
    >{\centering\arraybackslash}X
    >{\centering\arraybackslash}X
    >{\centering\arraybackslash}X
    >{\centering\arraybackslash}X
    >{\centering\arraybackslash}X
}

        \toprule
        \multirow{2.5}{*}{Method} & \multirow{2.5}{*}{$\lambda$} & \multicolumn{3}{c}{BLIP} & \multicolumn{3}{c}{XVLM} \\
        \cmidrule(lr){3-5}
        \cmidrule(lr){6-8}
        & & Text R@1 & Image R@1 & \multicolumn{1}{c}{Avg. R@1} & Text R@1 & Image R@1 & Avg. R@1\\
        \midrule

        \multirow{9}{*}{\textsc{chita}} & $10^{-5}$ & 0.00 & 0.02 & 0.01 & 0.50 & 0.34 & 0.42 \\ 
        & $10^{-4}$ & 0.00 & 0.02 & 0.01 & 0.80  & 0.71 & 0.76 \\ 
        & $10^{-3}$ & 0.02 & 0.02 & 0.02 & 11.48 & 9.23 & 10.36 \\ 
        & $10^{-2}$ & 62.10 & 47.63 & 54.87 & 64.98 & 50.71 & 57.85 \\ 
        & $10^{-1}$ & 66.42 & 50.85 & \textbf{58.64} & 72.54 & 57.45 & 65.00 \\ 
        & $10^0$ & 65.48 & 50.20 & 57.84 & 72.54 & 57.40 & 64.97 \\ 
        & $10^1$ & 66.44 & 50.77 & 58.61 & 72.78 & 57.30 & 65.04 \\ 
        & $10^2$ & 65.10 & 50.42 & 57.76 & 72.66 & 57.23 & 64.94 \\
        & $10^3$ & 65.42 & 50.04 & 57.73 & 72.84 & 57.33 & \textbf{65.09} \\ 
        \cmidrule(lr){1-8}
        
        \multirow{9}{*}{\textsc{chita++}} & $10^{-5}$ & 0.06 & 0.02 & 0.04 & 0.02 & 0.02 & 0.02 \\ 
        & $10^{-4}$ & 0.00 & 0.02 & 0.01 & 0.00 & 0.02 & 0.01 \\ 
        & $10^{-3}$ & 0.02 & 0.02 & 0.02 & 0.06 & 0.02 & 0.04 \\ 
        & $10^{-2}$ & 0.00 & 0.02 & 0.01 & 61.64 & 48.44 & 55.04 \\ 
        & $10^{-1}$ & 68.18 & 52.01 & \textbf{60.10} & 73.80 & 58.77 & \textbf{66.29} \\
        & $10^{0}$ & 67.26 & 51.85 & 59.56 & 72.96 & 58.22 & 65.59 \\ 
        & $10^{1}$ & 66.00 & 50.75 & 58.38 & 73.44 & 57.71 & 65.58 \\ 
        & $10^{2}$ & 65.64 & 50.32 & 57.98 & 73.12 & 57.58 & 65.35 \\ 
        & $10^{3}$ & 66.52 & 50.87 & 58.70 & 72.76 & 57.33 & 65.05 \\ 
        \bottomrule
        
    \end{tabularx}
    }
    \caption{Hyperparameter Grid Search on $\lambda$ (ridge penalty) for \textsc{chita} (top) and \textsc{chita++} (bottom) on both BLIP (left) and XVLM (right). The "Avg. R@1" column reports the average between "Text R@1" and "Image R@1", and is used for the final choice of $\lambda$.}
    \label{tab:chita-hp-search}
    
\end{table*}

\begin{table*}[!ht]
\centering
{\scriptsize
\begin{tabularx}{2\columnwidth}{
>{\centering\arraybackslash}p{1.5cm}
>{\centering\arraybackslash}p{0.75cm}
>{\centering\arraybackslash}p{1.10cm}
>{\centering\arraybackslash}p{0.75cm}
>{\centering\arraybackslash}X
>{\centering\arraybackslash}p{1.50cm}
>{\centering\arraybackslash}p{0.6cm}
>{\centering\arraybackslash}p{0.85cm}
>{\centering\arraybackslash}X
}

\toprule
Model & Task & Batch Size & Epochs & Optimizer & LR & WD & Scheduler & Warmup \\ 
\midrule

\multirow{3}{*}{BLIP} & ITR & 128 & 6 & AdamW, $\beta=(0.9, 0.999)$ & $10^{-5}$ & $0.05$ & cosine & \xmark \\ 
& IC & 256 & 5 & AdamW, $\beta=(0.9, 0.999)$ & $10^{-5}$ & $0.05$ & cosine & \xmark \\ 
& VQA & 256 & 10 & AdamW, $\beta=(0.9, 0.999)$ & $2 \times 10^{-5}$ & $0.05$ & cosine & \xmark \\ 
\cmidrule(lr){1-9}

\multirow{3}{*}{XVLM} & ITR & 128 & 10 & AdamW, $\beta=(0.9, 0.98)$ & $3\times10^{-5}$ & $0.01$ & linear & \checkmark, $10\%$ steps \\ 
& IC & 256 & 5 & AdamW, $\beta=(0.9, 0.98)$ & $10^{-5}$ & $0.01$ & linear & \checkmark, $10\%$ steps \\ 
& VQA & 256 & 10 & AdamW, $\beta=(0.9, 0.98)$ & $5 \times 10^{-5}$ & $0.01$ & linear & \checkmark, $10\%$ steps \\ 
\bottomrule

\end{tabularx}
}
\caption{Fine-tuning hyperparameter configurations for different tasks and VLMs.}
\label{tab:hps}
\end{table*}

\section{Limitations, Discussion and Future Works}\label{sec:limitations}
Here, we provide final considerations on \taskFull and \method. 

\myparagraph{Structured vs Unstructured sparsity.} The entire benchmarking of the main paper, as well as \method, are designed and evaluated with unstructured sparsity.
While this particular kind of sparsity already brings advantages in terms of memory reduction (\eg, via the standard Compressed Sparse Row format) and quantitative performance, it struggles in bringing actual advantages in terms of FLOPs/runtime reductions, given the current state of neural network training and GPU devices.
Nevertheless, advances in a parallel research line focusing on fast sparse operations have the potential to close this gap in upcoming years.
Notable examples of active research in this direction are sparse GPU kernels \cite{gale2020sparse} and NVIDIA's N:M sparsity acceleration.
Moreover, research in unstructured sparsity has proven beneficial as a starting point to design structured pruning algorithms, such as in \cite{molchanov2019importance}, where the importance of multiple connections is aggregated to obtain a neuron-level criterion, or \cite{michel2019sixteen}, where a similar idea is applied at the attention-head level in language transformers.
Consequently, since \method already integrates an explicit formulation of neuron-level importance, we believe extending it to structured sparsity is a promising direction for future research.

\myparagraph{Broader impact.} 
In our benchmark, we extensively study BLIP and XVLM in the context of \task, since they represent two significantly different ways to process and exploit multimodal information from visual and linguistic sources.
We believe our work can pave the way to explore pruning also Vision-Language systems employing different ways to cope with multimodal sources (\eg, \cite{chen2020uniter} or \cite{li2023blip2}), or to explore whether our findings also apply when other modalities, such as audio, are involved \cite{tang2023any}.
Finally, we believe that good algorithms for \taskFull can help reduce costs and energy requirements for both pruning and training Vision-Language systems in upcoming years, and do not see direct risks associated with our work.

\end{document}